\documentclass[letterpaper, 10 pt, conference]{ieeeconf}  

\IEEEoverridecommandlockouts                              

\usepackage{tikz}
\usepackage{booktabs}
\usepackage{multirow}
\usepackage{amsmath}
\usepackage{amssymb}
\usepackage{graphicx}
\usepackage{subcaption}
\usepackage{caption}
\usepackage[switch]{lineno}
\usepackage[margin=1in]{geometry}
\usepackage{array} 

\title{\LARGE \bf
Evaluating Histogram Matching for Robust Deep Learning–Based Grapevine Disease Detection
}

\author{Ruben~Pascual$^{1}$, Inés~Hernández$^{2}$, Salvador~Gutiérrez$^{3}$, Javier~Tardaguila$^{2}$, Pedro Melo-Pinto$^{4}$,\\Daniel~Paternain$^{1}$ and Mikel~Galar$^{1}$ 
\thanks{$^{1}$Institute of Smart Cities (ISC), Dept. of Statistics, Computer Science and Mathematics, Public Univ. of Navarre (UPNA), 31006 Pamplona, Spain {\tt\small \{ruben.pascual, daniel.paternain, mikel.galar\}@unavarra.es}}%
\thanks{$^{2}$Institute of Grapevine and Wine Sciences (Univ. of La Rioja, CSIC, Govt. of La Rioja), 26007, Logroño, Spain and Televitis Research Group, Univ. of La Rioja, 26006, Logroño, Spain {\tt\small \{ines.hernandez, javier.tardaguila\}@unirioja.es}}%
\thanks{$^{3}$Dept. of Computer Science and AI, Univ. of Granada, 18071, Granada, Spain {\tt\small salvaguti@decsai.ugr.es}}%
\thanks{$^{4}$Centre for the Research and Technology of Agroenvironmental and Biological Sciences (CITAB), Inov4Agro, and Departamento de Engenharias, UTAD, Quinta Dos Prados, 5000-801, Vila Real, Portugal {\tt\small pmelo@utad.pt}}%
}

\let\oldtwocolumn\twocolumn
\renewcommand\twocolumn[1][]{%
  \oldtwocolumn[{#1}{
    \vspace*{-1.25\baselineskip} 
    \begin{center}
    \includegraphics[width=\textwidth]{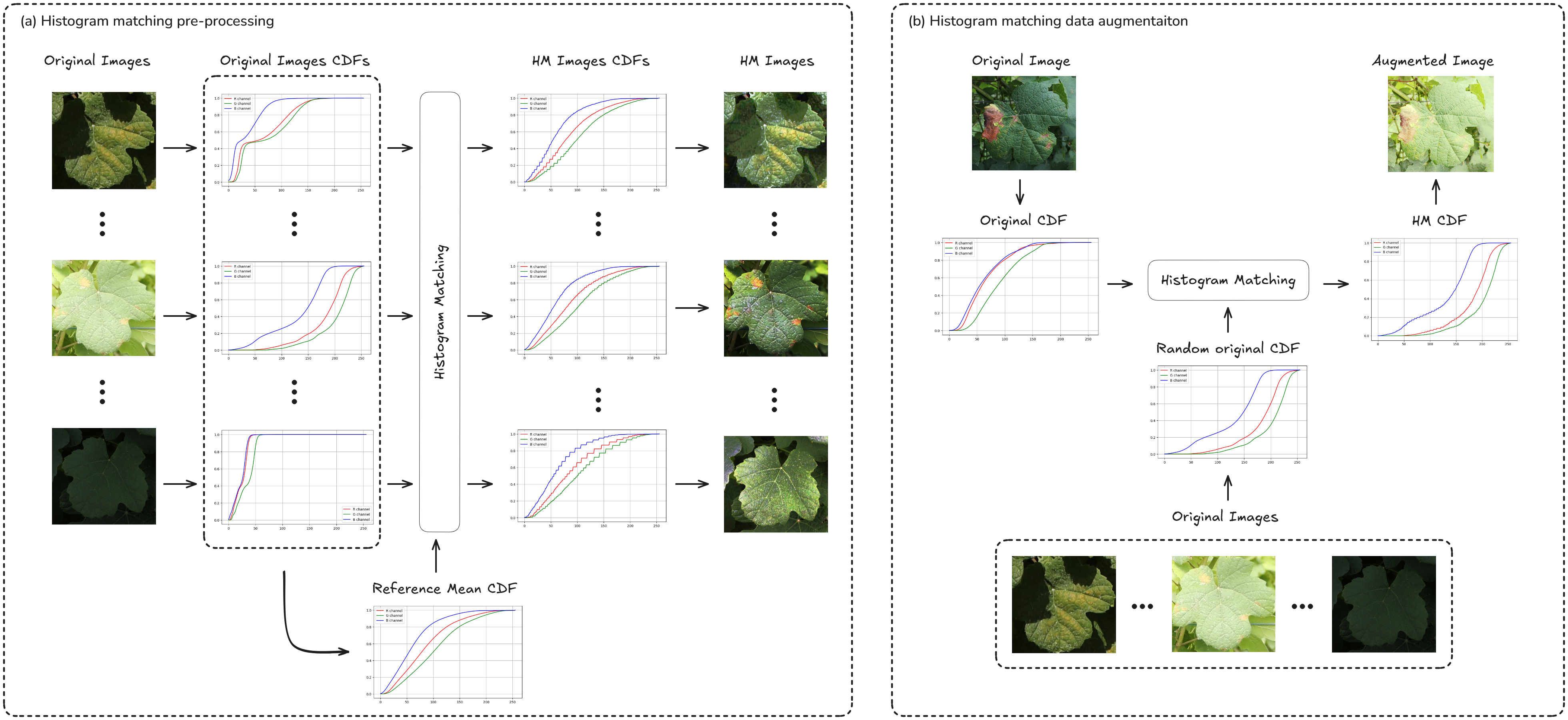}
    \captionof{figure}{\textbf{Illustration of the proposed histogram matching strategies.}
    (a) Histogram matching pre-processing: the cumulative distribution functions (CDFs) of all RGB channels are averaged across the dataset to obtain a reference mean CDF. Each image is then adjusted to match this reference.
    (b) Histogram matching data augmentation: during training, a random image from the original dataset is selected as reference, and histogram matching is applied between their CDFs to generate a color-augmented variant.}
    \label{fig:HM-diagram}
    \end{center}
  }]
}

\begin{document}
\bstctlcite{IEEEexample:BSTcontrol}

\maketitle
\thispagestyle{empty}
\pagestyle{empty}

\begin{abstract}

Variability in illumination is a primary factor limiting deep learning robustness for field-based plant disease detection. This study evaluates Histogram Matching (HM), a technique that transforms the pixel intensity distribution of an image to match a reference profile, to mitigate this in grapevine classification, distinguishing among healthy leaves, downy mildew, and spider mite damage. We propose a dual-stage integration of HM: (i) as a preprocessing step for normalization, and (ii) as a data augmentation technique to introduce controlled training variability. Experiments using 1,469 RGB images (comprising homogeneous leaf-focused and heterogeneous canopy samples) to train ResNet-18 models demonstrate that this combination significantly enhances robustness on real-world canopy images. While leaf-focused samples showed marginal gains, the canopy subset improved markedly, indicating that balancing normalization with histogram-based diversification effectively bridges the domain gap caused by uncontrolled lighting.

\end{abstract}

\section{INTRODUCTION}

Grapevine is one of the most widely cultivated and economically valuable fruit crops worldwide, with a cultivated area exceeding 7~million hectares in 2024~\cite{oiv2024report}. However, grapevine productivity and quality are susceptible to a wide variety of diseases and pests that, if not managed correctly and in a timely manner, can cause significant economic losses~\cite{Galet1996,wilcox2015}. Among them, downy mildew~\cite{unger2007CourseCTDVGPVICIHI} and some spider mites~\cite{wilcox2015} have been identified as prevalent stress factors in the vineyard. The differentiation of the causal agent in the field can be particularly challenging due to the similarities in their early visual symptoms, characterised by yellow spots on the adaxial side of the leaf surface~\cite{gutierrez2021DeepLDDMSMGFC}.

Traditionally, disease detection relies on visual inspection by field personnel, a subjective and labor-intensive process that complicates monitoring extensive cultivars~\cite{mokhtar2015SVMBasedDTLD,bock2020VisualEFASMPDSSCIA}. Moreover, definitive confirmation often demands destructive laboratory analyses, significantly increasing diagnosis time and operational costs~\cite{Gindro2014}. To address these limitations, computer vision has emerged as a rapid, non-destructive alternative, shifting toward deep learning architectures capable of extracting complex representations directly from raw data~\cite{ding2024NextGCVPDMPACSTEFD}. These models generalize effectively across agricultural environments~\cite{wang2025AdvancesDLAPDPDR}, with CNNs in viticulture successfully distinguishing downy mildew from spider mite damage~\cite{gutierrez2021DeepLDDMSMGFC} and object detection frameworks enabling precise lesion localization~\cite{pinheiro2023DeepLYSGBDABL}.

Despite these advances, model robustness is severely compromised by acquisition variability. Field images exhibit significant fluctuations in illumination and color that create a domain gap; seminal works demonstrated accuracy drops to as low as 33\% under reversed conditions~\cite{ferentinos2018DeepLMPDDD}, while recent benchmarks confirm that even state-of-the-art models degrade to 70–85\% in real-world environments~\cite{shafay2025RecentAPDDCO}. Factors such as background complexity and covariate shift further hinder reproducibility~\cite{barbedo2018FactorsIUDLPDR}. These findings underscore the critical need for preprocessing strategies capable of mitigating acquisition effects to ensure reliable field deployment.

To address these challenges, histogram matching offers a straightforward approach to reduce differences in brightness and color between images by aligning their intensity distributions to a reference~\cite{gonzalez2008DigitalIP}. Recent studies have revisited this classical technique from a deep learning perspective, showing that its careful integration can enhance image representation and improve classification accuracy~\cite{otsuka2025RethinkingIHMIC}. Such findings suggest that, when properly applied, histogram matching could help reduce domain shift and increase model robustness under varying illumination or sensor conditions.

Building on these insights, our objective is to assess whether histogram-based strategies can improve the generalization and robustness of CNNs when classifying grapevine leaf images affected by common stressors such as downy mildew and spider mite damage. To achieve this, we explore the potential of histogram matching when integrated at two stages of the machine learning workflow: (i) as a pre-processing step and (ii) as a data augmentation technique. To the best of our knowledge, this represents the first systematic evaluation of histogram matching applied across multiple stages of the training process for grapevine disease detection.

To validate the proposed approach, we conducted a rigorous study using a dataset of 1,469 RGB images, comprising both leaf-focused and canopy-derived samples classified into healthy, downy mildew, and spider mite categories. We trained a ResNet-18 architecture under varying input resolutions (128×128, 256×256, and 512×512) to systematically compare the impact of histogram matching when applied as a fixed preprocessing step versus a dynamic data augmentation strategy. All experiments were validated using a rigorous five-fold cross-validation scheme, repeated five times per configuration. Finally, model performance was assessed using Balanced Accuracy, ensuring a reliable evaluation that accounts for slight imbalances in class distribution.


\section{RELATED WORK}
\label{sec:related_work}

\vspace{6pt}\noindent\textbf{Deep learning for plant disease detection.} Deep learning revolutionized disease detection by automating feature extraction, with early CNNs significantly outperforming traditional methods~\cite{sladojevic2016DeepNNBRPDLIC}. Subsequent research introduced efficient architectures like VGG and ResNet to enhance accuracy~\cite{dey2022ComparativePFCDLVDHPTFDNDSROS,shafay2025RecentAPDDCO}, while Vision Transformers (ViTs) were later adopted to capture both fine details and global context~\cite{wu2021MultigranularityFEBVTTLDR}. Transfer learning has also been explored to address data scarcity by fine-tuning models pretrained on large-scale datasets for specific crops~\cite{morellos2022ComparisonDNNDFGDUTL}. In viticulture, models have successfully classified and segmented disease symptoms~\cite{hernandez2025EarlyDDMVUDNNSS,gutierrez2021DeepLDDMSMGFC}. However, maintaining robustness under real-world acquisition variability remains a critical challenge.

\vspace{6pt}\noindent\textbf{Challenges from illumination and acquisition variability.}
Variability in image acquisition conditions is one of the main factors limiting the performance and generalisation of deep learning models in plant phenotyping~\cite{barbedo2018FactorsIUDLPDR}. Natural lighting changes, shadows, and differences in camera sensors can introduce significant inconsistencies in colour and brightness, leading to shifts between training and deployment domains~\cite{ferentinos2018DeepLMPDDD}. Such discrepancies often cause deep models to overfit to dataset-specific visual patterns rather than learning disease-relevant features, resulting in degraded accuracy when tested on images captured under different conditions~\cite{mohanty2016UsingDLIPDD,shafay2025RecentAPDDCO}. Some works proposed standardising acquisition protocols to minimise background and illumination variability~\cite{barbedo2016ReviewMCAPDIBVRI}. However, these methods often require additional calibration data or introduce computational overhead. This limitation highlights the critical need for flexible, data-centric strategies that can normalize appearance without constraining the image capture process, shifting the focus toward solutions that operate directly within the learning pipeline.

\vspace{6pt}\noindent\textbf{Preprocessing and data augmentation for robustness.} To address the acquisition challenges described above, preprocessing techniques serve as a primary computational step, aiming to enhance dataset consistency by correcting illumination and color variations prior to model training. Classical image processing strategies, including histogram equalization and histogram matching, serve as foundational methods for normalizing pixel intensity distributions while preserving key visual structures~\cite{gonzalez2008DigitalIP}. These histogram-based approaches have been widely adapted to harmonize image appearance across diverse domains, as demonstrated in remote sensing~\cite{galar2020SuperResolutionSIUCNNRGTD} and image restoration tasks~\cite{sun2025RestoringIAWCHT, otsuka2025RethinkingIHMIC}.

While preprocessing focuses on standardization, data augmentation plays a complementary role by artificially increasing the diversity of training samples to improve generalization~\cite{shorten2019SurveyIDADL}. Common strategies utilize geometric transformations (e.g., rotation, flipping) for spatial invariance and photometric adjustments to enhance illumination robustness~\cite{taylor2018ImprovingDLGDA}. Furthermore, advanced techniques like sample pairing~\cite{inoue2018DataAPSIC} and random erasing~\cite{zhong2020RandomEDA} reduce overfitting by encouraging feature invariance to occlusion and mixed content. Beyond handcrafted transformations, generative models have been explored as a means of synthesizing new, realistic samples to enrich training datasets; for example, GAN-based augmentation has been shown to improve performance in data-scarce settings across several vision tasks~\cite{bowles2018GANAATDUGAN}.


\section{MATERIALS AND METHODS}
\label{sec:materials_methods}

Building on the insights presented in the related work, this study explores the dual integration of histogram matching (HM) as both a preprocessing and a data augmentation technique. To evaluate this approach, the methodology is structured into three stages. The first stage describes the dataset composition, distinguishing between leaf-focused and canopy-derived image subsets (Section \ref{sec:dataacq}). The second stage details the generation of comparative datasets using HM as a preprocessing step to normalize illumination and color variability (Section \ref{sec:prepro}). Finally, the third stage addresses model training and evaluation (Section \ref{sec:modeltrain}), comparing performance with and without the proposed HM data augmentation to assess robustness across all configurations.

\subsection{Dataset description}
\label{sec:dataacq}
\begin{figure}[ht]
    \centering
    \resizebox{\linewidth}{!}{%
        \begin{tikzpicture}[picture format/.style={inner sep=0.1pt,}]
            
              \node[picture format, outer sep=1] (O1) {\includegraphics[width=1in, height=1in]{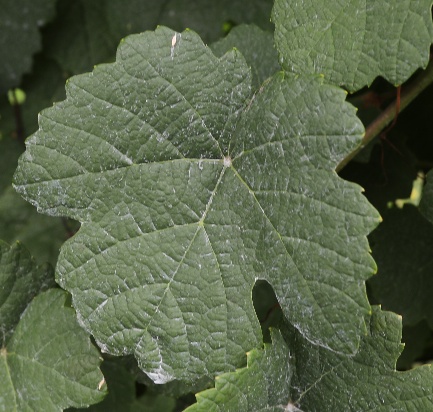}}; 
              \node[picture format,anchor=north west, outer sep=1] (O2) at (O1.north east) {\includegraphics[width=1in, height=1in]{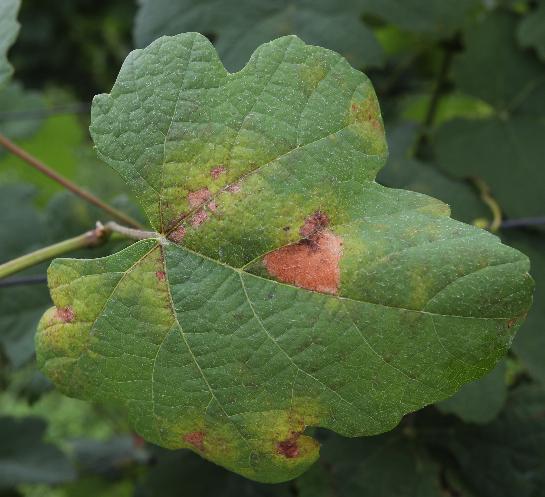}}; 
              \node[picture format,anchor=north west, outer sep=1] (O3) at (O2.north east) {\includegraphics[width=1in, height=1in]{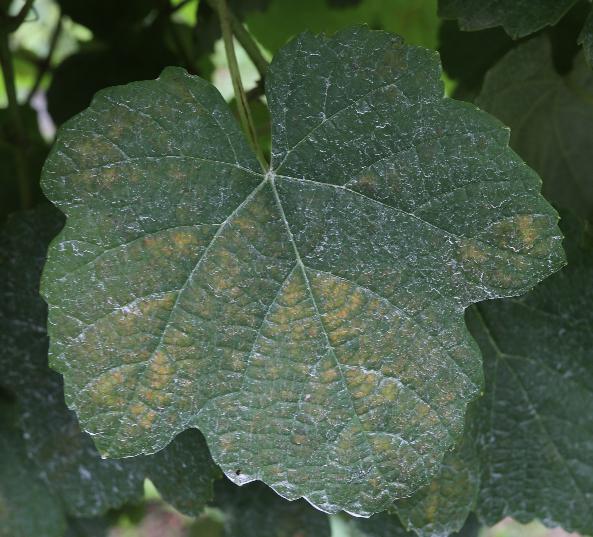}};
    
              \node[picture format,anchor=north, outer sep=1] (P1) at (O1.south){\includegraphics[width=1in, height=1in]{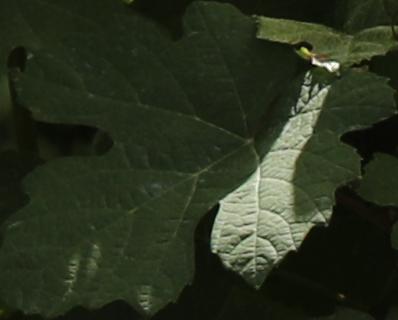}}; 
              \node[picture format,anchor=north west, outer sep=1] (P2) at (P1.north east) {\includegraphics[width=1in, height=1in]{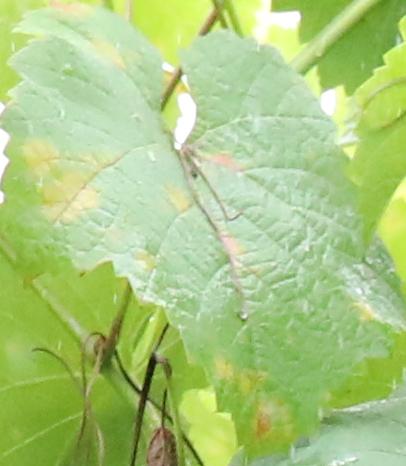}}; 
              \node[picture format,anchor=north west, outer sep=1] (P3) at (P2.north east) {\includegraphics[width=1in, height=1in]{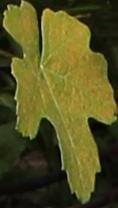}};
        
              \node[anchor=east, text width=0.9in] (T1) at (O1.west) {\large a) Leaf Focused};
              \node[anchor=east, text width=0.9in] (T2) at (P1.west) {\large b) Extracted from canopy images};
              \node[anchor=south] (H1) at (O1.north) {\large Healthy};
              \node[anchor=south] (H2) at (O2.north) {\large Downy mildew};
              \node[anchor=south] (H3) at (O3.north) {\large Spider mite};
          
        \end{tikzpicture}
    }
    \caption{Examples of leaf images from each image type and class used for the experiment.}
    \label{fig:leaf-images-examples}
\end{figure}

The dataset comprises 1,469 RGB images acquired in three commercial vineyards (Basque Country, Spain) during the 2019 and 2021 seasons, as detailed in~\cite{gutierrez2021DeepLDDMSMGFC}. The data is categorized into two subsets with distinct acquisition characteristics (Fig.~\ref{fig:leaf-images-examples}). Leaf-focused images were acquired manually, centering on the leaf to minimize background distraction and maximize visual clarity. In contrast, canopy-derived images were obtained from on-the-go platform monitoring, introducing realistic  heterogeneity regarding lighting, background complexity, and leaf orientation. For this specific subset, individual leaves were annotated and extracted using LabelImg~\cite{Tzutalin2015}. Table~\ref{tab:dataset} summarizes the final distribution.

\begin{table}[ht]
\caption{Number of images considered from each image type and class.}
\label{tab:dataset}
\resizebox{\linewidth}{!}{%
\begin{tabular}{@{}lrrrr@{}}
\toprule
Image type & Healthy & Downy mildew & Spider mite & Total \\ \midrule
Leaf-focused & 308 & 275 & 258 & 841  \\
Extracted from canopy images & 239 & 221 & 168 & 628 \\ \midrule
Both & 547 & 496 & 426  & 1469 \\ \bottomrule
\end{tabular}%
}
\end{table}

\subsection{Data pre-processing}
\label{sec:prepro}

\begin{figure}[ht]
    \centering
    \resizebox{\linewidth}{!}{%
        \begin{tikzpicture}[picture format/.style={inner sep=0.1pt,}]
            
              \node[picture format, outer sep=1] (O1) {\includegraphics[width=1in, height=1in]{imgs/sano_c.jpg}}; 
              \node[picture format,anchor=north west, outer sep=1] (O2) at (O1.north east) {\includegraphics[width=1in, height=1in]{imgs/mildiu_c.jpg}}; 
              \node[picture format,anchor=north west, outer sep=1] (O3) at (O2.north east) {\includegraphics[width=1in, height=1in]{imgs/arana_c.jpg}};
    
              \node[picture format,anchor=north, outer sep=1] (P1) at (O1.south){\includegraphics[width=1in, height=1in]{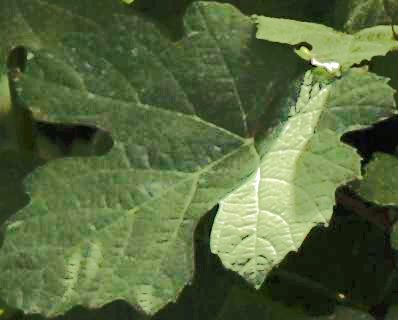}}; 
              \node[picture format,anchor=north west, outer sep=1] (P2) at (P1.north east) {\includegraphics[width=1in, height=1in]{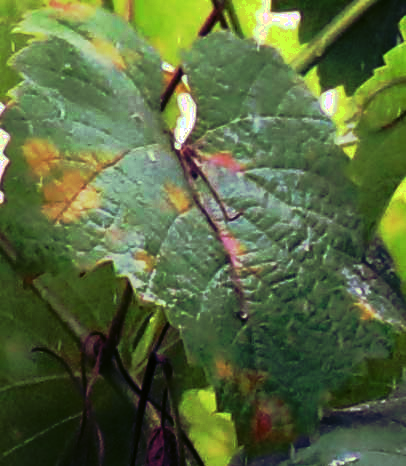}}; 
              \node[picture format,anchor=north west, outer sep=1] (P3) at (P2.north east) {\includegraphics[width=1in, height=1in]{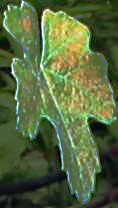}};
        
              \node[anchor=east, yshift=0.3cm, text width=0.9in, align=center] (T1) at (O1.west) {\large Original};
              \node[anchor=east, yshift=0.3cm, text width=0.9in, align=center] (T2) at (P1.west) {\large HM Processed};
              \node[anchor=south] (H1) at (O1.north) {\large Healthy};
              \node[anchor=south] (H2) at (O2.north) {\large Downy mildew};
              \node[anchor=south] (H3) at (O3.north) {\large Spider mite};
          
        \end{tikzpicture}
    }
    \caption{Examples of images before and after histogram matching pre-processing for all classes.}
    \label{fig:prepro_example}
\end{figure}

Image preprocessing aimed to homogenize illumination and color variations across the dataset while preserving the visual characteristics relevant for disease identification. The overall workflow is illustrated in Fig.~\ref{fig:HM-diagram}a. A modified version of histogram matching was implemented. Unlike the classical approach, in which the intensity distribution of a single image is matched to that of a chosen reference, our method estimates an average intensity distribution representative of the entire dataset. Specifically, normalized histograms were computed for each RGB channel over all images, and their mean distribution was used as a reference for matching. The cumulative distribution functions (CDFs) corresponding to both the image and reference histograms were then employed to perform the mapping between intensity levels, ensuring that each image’s channel-wise CDF closely matched the reference CDF. This resulted in a set of color-normalized images with reduced variability in brightness and illumination. An example of the processed images can be seen in Fig \ref{fig:prepro_example}. This procedure produced two datasets for subsequent experiments: one composed of the original images and another containing the histogram-matched versions (HM).

\subsection{Model training and data augmentation}
\label{sec:modeltrain}

The classification task was performed using a ResNet-18 convolutional neural network, selected for its balance between computational simplicity and strong performance. The network was initialized with ImageNet-pretrained weights and fine-tuned using the FastAI~2.8.4 framework. All trainings employed standard FastAI data augmentations, which include random flips, rotations, and lighting variations.

In addition to the default FastAI augmentations, a custom histogram-matching augmentation was introduced to inject controlled variability in color and illumination (Fig.~\ref{fig:HM-diagram}b). In this process, each training image had a 50\% probability of being paired with a randomly selected reference image from the original (non-histogram-matched) dataset. Histogram matching was then performed independently for each RGB channel by aligning the cumulative distribution functions (CDFs) of the training and reference images, generating a color-augmented variant with adjusted brightness and hue characteristics. This approach reintroduced moderate color variability into the training process, complementing the color normalization achieved during pre-processing and reducing potential overfitting to specific illumination conditions.

To systematically assess the influence of histogram matching when applied at different stages of the pipeline, experiments covered all combinations of input resolution ($128\times128$, $256\times256$, $512\times512$), preprocessing (with or without histogram matching), and data augmentation (with or without the proposed method). 

Given the variability observed in the dataset, particularly in canopy-derived images, a five-fold cross-validation strategy was implemented to obtain a robust estimate of model performance. In each iteration, 20\% of the data were reserved for testing, while the remaining samples were used for training. The folds were generated to maintain a balanced distribution of the three symptom categories and the two image types (leaf-focused and canopy-derived). This procedure ensured that every image was used for testing exactly once, providing a reliable evaluation while minimizing the influence of dataset partitioning on the final results.

Each configuration was evaluated across the five cross-validation folds, with each fold repeated five times to ensure statistical reliability. This resulted in a total of 300 trained models ($3~\text{resolutions} \times 2~\text{preprocessing} \times 2~\text{augmentation} \times 5~\text{folds} \times 5~\text{repetitions}$).

Model optimization was carried out using the Adam optimizer with a learning rate of $1\times10^{-3}$, a batch size of 64, and a maximum of 20~epochs, retaining the model with the lowest validation loss to prevent overfitting. All experiments were executed on a workstation equipped with an NVIDIA~RTX~A6000~GPU (48~GB VRAM), an AMD EPYC 9334 32-core processor, and 64~GB of RAM, running CUDA~13.0.


\section{RESULTS AND DISCUSSION}
\label{sec:results}

This section presents the classification results across all experimental configurations designed to evaluate the impact of histogram matching. Model performance was evaluated using balanced accuracy and per-class accuracy as complementary metrics. Balanced accuracy provides an overall measure that compensates for slight class imbalance by averaging recall across categories, while per-class accuracy offers a more detailed view of the model’s behavior for each symptom type, highlighting potential differences in sensitivity across classes.

To better understand the influence of image characteristics on model performance, the results were analyzed separately for the two subsets that compose the dataset: leaf-focused and canopy-derived images. The canopy-derived images exhibit greater variability in illumination, background, and leaf positioning, providing a closer representation of real field conditions but also a more challenging classification scenario. Consequently, while both subsets are presented for comparison, the canopy-based evaluation is considered the main reference for assessing the proposed methods.

Table~\ref{tab:canopy_leaf_combined} summarizes the classification performance across all experimental configurations and input resolutions for both canopy-derived and leaf-focused images. The table reports the mean and standard deviation of balanced accuracy and per-class accuracy over five repetitions and five cross-validation folds. Results are grouped by input size, dataset version (original or histogram-matched), and the use or absence of histogram-based data augmentation.

\begin{table*}[ht]
\centering
\small
\caption{Summary of classification results for canopy-derived and leaf-focused images, expressed as balanced accuracy and per-class accuracy (mean $\pm$ standard deviation across five repetitions and five cross-validation folds). The highest value for each input size is highlighted in \textbf{bold}.}
\resizebox{\textwidth}{!}{
\begin{tabular}{lllcccccccc}
\toprule
 &  &  & \multicolumn{4}{c}{\textbf{Canopy-derived images}} & \multicolumn{4}{c}{\textbf{Leaf-focused images}} \\
\cmidrule(lr){4-7} \cmidrule(lr){8-11}
\textbf{Size} & \textbf{Dataset} & \textbf{Aug.} & Balanced acc. & Spider mite & Downey mildew & Healthy & Balanced acc. & Spider mite & Downey mildew & Healthy \\
\midrule
128 & Original & No & $.8245 {\scriptscriptstyle \pm .0345}$ & $.8298 {\scriptscriptstyle \pm .0623}$ & $.7763 {\scriptscriptstyle \pm .0562}$ & $.8673 {\scriptscriptstyle \pm .0605}$ & $\pmb{.9214 {\scriptscriptstyle \pm .0190}}$ & $.9296 {\scriptscriptstyle \pm .0367}$ & $\pmb{.8545 {\scriptscriptstyle \pm .0662}}$ & $\pmb{.9799 {\scriptscriptstyle \pm .0183}}$ \\
 &  & Yes & $.8249 {\scriptscriptstyle \pm .0367}$ & $.8174 {\scriptscriptstyle \pm .0701}$ & $.7574 {\scriptscriptstyle \pm .0538}$ & $\pmb{.8998 {\scriptscriptstyle \pm .0478}}$ & $.9184 {\scriptscriptstyle \pm .0186}$ & $\pmb{.9382 {\scriptscriptstyle \pm .0307}}$ & $.8429 {\scriptscriptstyle \pm .0708}$ & $.9741 {\scriptscriptstyle \pm .0248}$ \\
 & HM & No & $.8387 {\scriptscriptstyle \pm .0276}$ & $.8551 {\scriptscriptstyle \pm .0511}$ & $.7687 {\scriptscriptstyle \pm .0631}$ & $.8924 {\scriptscriptstyle \pm .0300}$ & $.9083 {\scriptscriptstyle \pm .0277}$ & $.9100 {\scriptscriptstyle \pm .0395}$ & $.8473 {\scriptscriptstyle \pm .0492}$ & $.9676 {\scriptscriptstyle \pm .0206}$ \\
 &  & Yes & $\pmb{.8470 {\scriptscriptstyle \pm .0274}}$ & $\pmb{.8557 {\scriptscriptstyle \pm .0538}}$ & $\pmb{.8009 {\scriptscriptstyle \pm .0533}}$ & $.8845 {\scriptscriptstyle \pm .0345}$ & $.9053 {\scriptscriptstyle \pm .0296}$ & $.8946 {\scriptscriptstyle \pm .0491}$ & $.8524 {\scriptscriptstyle \pm .0528}$ & $.9689 {\scriptscriptstyle \pm .0233}$ \\
\midrule
256 & Original & No & $.8630 {\scriptscriptstyle \pm .0333}$ & $.8428 {\scriptscriptstyle \pm .0595}$ & $.8673 {\scriptscriptstyle \pm .0408}$ & $.8790 {\scriptscriptstyle \pm .0416}$ & $.9596 {\scriptscriptstyle \pm .0183}$ & $.9504 {\scriptscriptstyle \pm .0266}$ & $.9447 {\scriptscriptstyle \pm .0408}$ & $.9837 {\scriptscriptstyle \pm .0156}$ \\
 &  & Yes & $.8711 {\scriptscriptstyle \pm .0351}$ & $.8358 {\scriptscriptstyle \pm .0602}$ & $.8655 {\scriptscriptstyle \pm .0561}$ & $.9121 {\scriptscriptstyle \pm .0472}$ & $\pmb{.9618 {\scriptscriptstyle \pm .0155}}$ & $\pmb{.9543 {\scriptscriptstyle \pm .0229}}$ & $\pmb{.9455 {\scriptscriptstyle \pm .0328}}$ & $.9856 {\scriptscriptstyle \pm .0128}$ \\
 & HM & No & $.8842 {\scriptscriptstyle \pm .0244}$ & $\pmb{.8672 {\scriptscriptstyle \pm .0662}}$ & $.8736 {\scriptscriptstyle \pm .0444}$ & $.9117 {\scriptscriptstyle \pm .0454}$ & $.9457 {\scriptscriptstyle \pm .0265}$ & $.9431 {\scriptscriptstyle \pm .0421}$ & $.9142 {\scriptscriptstyle \pm .0326}$ & $.9798 {\scriptscriptstyle \pm .0272}$ \\
 &  & Yes & $\pmb{.8949 {\scriptscriptstyle \pm .0271}}$ & $.8643 {\scriptscriptstyle \pm .0550}$ & $\pmb{.8764 {\scriptscriptstyle \pm .0549}}$ & $\pmb{.9440 {\scriptscriptstyle \pm .0280}}$ & $.9584 {\scriptscriptstyle \pm .0186}$ & $.9531 {\scriptscriptstyle \pm .0314}$ & $.9324 {\scriptscriptstyle \pm .0377}$ & $\pmb{.9896 {\scriptscriptstyle \pm .0123}}$ \\
\midrule
512 & Original & No & $.8796 {\scriptscriptstyle \pm .0293}$ & $.8523 {\scriptscriptstyle \pm .0605}$ & $.8855 {\scriptscriptstyle \pm .0492}$ & $.9009 {\scriptscriptstyle \pm .0449}$ & $.9685 {\scriptscriptstyle \pm .0196}$ & $.9544 {\scriptscriptstyle \pm .0451}$ & $.9622 {\scriptscriptstyle \pm .0277}$ & $\pmb{.9890 {\scriptscriptstyle \pm .0130}}$ \\
 &  & Yes & $.8998 {\scriptscriptstyle \pm .0247}$ & $.8756 {\scriptscriptstyle \pm .0691}$ & $\pmb{.8945 {\scriptscriptstyle \pm .0506}}$ & $.9293 {\scriptscriptstyle \pm .0349}$ & $\pmb{.9712 {\scriptscriptstyle \pm .0146}}$ & $\pmb{.9606 {\scriptscriptstyle \pm .0374}}$ & $\pmb{.9680 {\scriptscriptstyle \pm .0205}}$ & $.9851 {\scriptscriptstyle \pm .0162}$ \\
 & HM & No & $.8987 {\scriptscriptstyle \pm .0252}$ & $.8861 {\scriptscriptstyle \pm .0578}$ & $.8806 {\scriptscriptstyle \pm .0481}$ & $.9292 {\scriptscriptstyle \pm .0317}$ & $.9559 {\scriptscriptstyle \pm .0173}$ & $.9362 {\scriptscriptstyle \pm .0411}$ & $.9491 {\scriptscriptstyle \pm .0273}$ & $.9826 {\scriptscriptstyle \pm .0147}$ \\
 &  & Yes & $\pmb{.9055 {\scriptscriptstyle \pm .0227}}$ & $\pmb{.8890 {\scriptscriptstyle \pm .0510}}$ & $.8861 {\scriptscriptstyle \pm .0594}$ & $\pmb{.9414 {\scriptscriptstyle \pm .0271}}$ & $.9701 {\scriptscriptstyle \pm .0144}$ & $.9592 {\scriptscriptstyle \pm .0301}$ & $.9658 {\scriptscriptstyle \pm .0231}$ & $.9852 {\scriptscriptstyle \pm .0174}$ \\
\bottomrule
\end{tabular}
}
\label{tab:canopy_leaf_combined}
\end{table*}

Overall, the results indicate that histogram matching contributed to consistent performance improvements across most configurations. Across the board, increasing input resolution enhanced classification accuracy, although performance gains became marginal beyond $256\times256$ pixels, suggesting that the model captures the majority of discriminative features at moderate resolutions. The gains from histogram matching were most pronounced for the canopy-derived images, where lighting and color variability were inherently higher. For this subset, the combination of histogram-matched preprocessing and histogram-based augmentation yielded the highest balanced accuracy, reaching up to 0.9055 for $512\times512$ inputs.

The observed benefits in these heterogeneous conditions stems from the dual role that histogram matching plays in the pipeline. First, when applied as a preprocessing step, it acts as a standardization mechanism, mitigating inconsistencies in brightness and color balance by aligning image histograms to a common reference. Second, when used as a data augmentation strategy, it acts in the opposite direction, deliberately introducing controlled variations in color distributions. This variability prevents overfitting to specific lighting conditions and encourages the model to focus on invariant cues such as texture and lesion morphology. By balancing uniformity (preprocessing) and variability (augmentation), the model achieves greater robustness in scenarios where illumination fluctuations are unavoidable.

In contrast, for the leaf-focused subset, where image variability was considerably lower, the impact of histogram matching was less consistent. At lower resolutions ($128\times128$), both preprocessing and augmentation slightly reduced accuracy compared to the original dataset. This suggests that in homogeneous images, where leaf position and background remain stable, color differences play a discriminative role; thus, excessive normalization may inadvertently remove informative cues. However, at higher resolutions ($512\times512$), the histogram-matching augmentation provided modest improvements, confirming that while normalization must be applied cautiously in stable environments, the data augmentation remains valuable for generalization.

From a practical perspective, the substantial improvements achieved on canopy-derived images are more relevant than the performance observed in the leaf-focused subset, as the former significantly better represent natural field conditions. In production environments, scalable disease monitoring relies on non-destructive, on-the-go imaging of the vine canopy rather than the manual acquisition of individual leaves under controlled lighting. Consequently, the ability of histogram matching to stabilize performance in these complex, unconstrained scenarios is the primary indicator of its real-world utility. This resilience highlights the relevance of the approach for viticulture monitoring systems, particularly those relying on low-cost consumer-grade RGB cameras. Since canopy-level images inherently contain strong lighting gradients and complex backgrounds, the ability to enhance model stability without the need for specialized sensors or controlled acquisition setups represents a valuable step toward scalable, real-world disease detection.

Despite these promising outcomes, some limitations must be acknowledged. The dataset included only two grapevine cultivars and two types of stress symptoms, which may limit generalization to other varieties or conditions. Moreover, histogram matching primarily corrects global illumination differences, while local shading or occlusion effects remain challenging. Future work should explore adaptive or learnable histogram matching methods~\cite{otsuka2025RethinkingIHMIC} to capture more localized color adjustments. Testing on other crops and integrating the approach into real-time phenotyping systems would further validate its utility for precision viticulture.

\section{CONCLUSION}
\label{sec:conclusion}

This study presented a systematic evaluation of histogram matching for grapevine disease detection, examining its role as both a preprocessing normalization step and a data augmentation strategy across multiple input resolutions. The results show that histogram matching consistently improves robustness, with the largest gains observed for canopy-derived images, those most representative of real field conditions, while leaf-focused images, acquired under more homogeneous settings, already achieve high accuracy. Using histogram matching in both stages proved complementary: preprocessing reduced illumination and color-induced variability, and augmentation reintroduced controlled diversity that encouraged invariance to acquisition conditions.

Taken together, these findings support histogram matching as a simple, computationally efficient tool to enhance deep-learning pipelines for vineyard monitoring with standard RGB imagery. Future work should investigate adaptive or learnable variants of histogram matching, extend validation to additional cultivars and stress types, and benchmark the approach against alternative normalization and domain adaptation strategies in larger, multi-site datasets and real-time field monitoring scenarios to further test scalability and generalization.

\section*{Acknowledgments}

This work was supported by the Spanish Ministry of Science and Innovation under project PID2022-\hspace{0pt}136627NB-\hspace{0pt}I00 (MCIN\hspace{0pt}/AEI\hspace{0pt}/10.13039\hspace{0pt}/501100011033\hspace{0pt}/\hspace{0pt}FEDER, UE) and by the European Union Horizon 2020 FET Open program under grant agreement ID 828940 (Project NoPest). Regarding individual support, Rubén Pascual was funded by an FPU grant (FPU22/02961), and Inés Hernández received the FPI PhD grant (1150/2020) from Universidad de La Rioja and Gobierno de La Rioja. Finally, the authors acknowledge the use of Google's Gemini for assistance with language editing and proofreading.

\bibliographystyle{IEEEtran}
\bibliography{authorControl,Leafs}

\end{document}